\documentclass[10pt,twocolumn,letterpaper]{article}
\pdfoutput=1

\usepackage{epsfig}
\usepackage{graphicx}
\usepackage{amsmath}
\usepackage{amssymb}

\usepackage{amsthm}
\usepackage{paralist}
\usepackage{wasysym}
\usepackage{microtype}
\usepackage{subfigure}
\usepackage{amsthm}
\usepackage{epsfig}
\usepackage{graphicx}
\usepackage[utf8]{inputenc}
\usepackage{color}

\usepackage{ifthen}
\newboolean{extended}
\setboolean{extended}{false}
\usepackage{wrapfig}
\usepackage{ifthen}
\newcommand{\db}{\ensuremath{D}}

\setboolean{extended}{false}

\usepackage[pagebackref=true,breaklinks=true,letterpaper=true,colorlinks,bookmarks=false]{hyperref}

\date{}

\begin{document}

\title{Minimizing the Number of Matching Queries\\ for Object Retrieval}

\author{Johannes Niedermayer, Peer Kröger\\
LMU Munich\\
Germany\\
{\tt\small {niedermayer,kroeger}@dbs.ifi.lmu.de}
}


\maketitle

\begin{abstract}
To increase the computational efficiency of interest-point based object retrieval, researchers have put remarkable research efforts into improving the
efficiency of $k$NN-based feature matching, pursuing to match thousands of features against a database within fractions of a second. However, due to
the high-dimensional nature of image features that reduces the effectivity of index structures (curse of dimensionality), due to the vast amount of
features stored in image databases (images are often represented by up to several thousand features), this ultimate goal demanded to trade query
runtimes for query precision. In this paper we address an approach complementary to indexing in order to improve the runtimes of retrieval by querying
only the most promising keypoint descriptors, as this affects matching runtimes linearly and can therefore lead to increased efficiency. As this reduction of
$k$NN queries reduces the number of tentative correspondences, a loss of query precision is minimized by an additional image-level correspondence
generation stage with a computational performance independent of the underlying indexing structure. We evaluate such an adaption of the standard recognition pipeline
on a variety of datasets using both SIFT and state-of-the-art binary descriptors. Our results suggest that decreasing the number of queried
descriptors does not necessarily imply a reduction in the result quality as long as alternative ways of increasing query recall (by thoroughly
selecting $k$) and MAP (using image-level correspondence generation) are considered.
\end{abstract}

\section{Introduction}
While the development of the SIFT-Descriptor \cite{Low04} made effective object retrieval on a large scale feasible, its initial use of nearest
neighbor queries lead to slow runtimes even on relatively small data sets. In 2003, the invention of the Bag of Visual Words (BoVW) technique
\cite{SivZis03} aimed at solving this issue by roughly approximating the matching step using quantization, initiating a whole new area of research.
However soon the limitations of this rough approximation became obvious, enforcing the development of more accurate techniques for assigning query
vectors to database features. Whilst initial approaches aiming at increasing the accuracy of the matching step such as soft assignment
\cite{PhiChuIsaSivZis08} were relatively close to the BoVW approach, the focus in recent years turned back more and more to approximate $k$NN queries
\cite{KalAvr14,GeHeKeSun13,BabLem12,NorPunFle12} due to their possible gain in matching accuracy \cite{JegDouSch11b}: $k$NN queries provide
an accurate ranking of the matching candidates and a measure of proximity between feature vectors and query vectors. This additional information can
be exploited for weighting the scores of image matches, increasing retrieval accuracy considerably \cite{JegDouSch11b}.

Current research on $k$NN processing in the image retrieval community focuses on maximizing accuracy, on minimizing the memory footprint of index
structure and feature vectors, and on minimizing processing time. In recent years such techniques have received a vast amount of interest even in the
most prestigious conferences addressing image retrieval\cite{KalAvr14,GeHeKeSun13,BabLem12,JegDouSch11a,NorPunFle12}.
As a result, a remarkable leap in performance has been achieved concerning efficient and effective $k$NN query processing. However, with the vast
amount of features that have to be matched during recognition (up to a few thousand), even very fast $k$NN indexing techniques that can
provide approximate query results in under ten milliseconds (e.g. \cite{KalAvr14}), would yield recognition runtimes of many seconds.

We argue that the use of $k$NN queries for object recognition in large-scale systems cannot be achieved by developing efficient indexing techniques
alone. The problem of efficiency has to be approached from different research directions as well, such as the \textit{number} of $k$NN queries posed
on the system, as reducing the number of $k$NN queries linearly decreases the runtime of the matching step.
In this paper we aim at addressing this problem. We evaluate an alternative recognition pipeline that ranks features extracted from the query image by
assessing their matchability. Then, the most promising features in this ranking are matched against the database using traditional $k$NN queries.
However, despite gaining efficiency, the enforced reduction of $k$NN queries causes a reduction of feature matches, decreasing the quality of the
query result. While recall can be increased by increasing $k$, to increase Mean Average Precision (MAP) we propose to expand matches on the image
level:
Given a single seed feature match in a candidate image, this match is expanded by comparing its spatially neighboring keypoints. The idea of this
additional step is to push load from the matching step (with complexity mostly determined by the underlying index structure) to an additional step
that only has to consider the features stored in an image pair. The resulting enriched set of matches can then be processed equivalently to
techniques based on BoVW, e.g. by using query expansion \cite{ChuPhiSivIsaZis07} or geometric verification \cite{PhiChuIsaSivZis07}.

This work stands in contrast to research in the area of BoVW-based retrieval: Research involving the BoVW pipeline often assumes that the
matching step is relatively cheap, especially if approximate cluster assignment techniques such as hierarchical $k$-means \cite{NisSte06} or
approximate $k$-means \cite{PhiChuIsaSivZis07} are used. Therefore such research focused on increasing Mean Average Precision (MAP) at a large
number of query features.
In contrast, this paper aims at maximizing MAP for a small number of processed features. This different optimization criterion is
especially of interest as techniques that do not lead to significant gains in performance at a high number of features (where convergence to the
maximum possible MAP has already been achieved by other techniques) can lead to a remarkably higher MAP when only a low number of features is
queried. 

The contribution of this paper is to provide a \textit{simple} and \textit{extensible} pipeline for \textit{large-scale} object retrieval based on
\textit{$k$NN queries} with \textit{all} of the following properties:
\begin{compactitem}
  \item \textit{Reduction of the number of keypoints queried} by a general keypoint ranking scheme in order to reduce matching times. The pipeline is
  not bound to a specific keypoint selection technique as long as keypoints can be ranked by their estimated quality.
  \item Acceleration of the pipeline by state-of-the art index structures such as (Locally Optimized) Product quantization \cite{JegDouSch11a} or
  Multi-Index-Hashing\cite{NorPunFle12}.
  \item \textit{Geometric Match Expansion} to relieve the index structure and to increase query \textit{MAP}.
  \item The use of many nearest neighbors ($k > 2$) to increase the number of seed hypotheses and therefore query \textit{recall}.
  \item Consideration of \textit{distances} between features during score generation to allow accurate scoring of image features by their similarity.
\end{compactitem}
We further provide a thorough evaluation of this pipeline on a variety of well-known datasets, including Oxford5k, Oxford105k, Paris 6k, and
INRIA Holidays, provide insights into advantages and disadvantages of the approach, and show that such match expansion techniques can lead to
performance improvements. We also evaluate the effect of $k$ in relation to the number of keypoints queried on the systems performance, and the
pipeline's behaviour on different feature descriptors including real-valued (SIFT) and binary (BinBoost) features.

This paper is organized as follows. Section \ref{sec:problemdef} formally defines the problem addressed in this paper. We then review related work in
Section~\ref{sec:related}. In Section~\ref{sec:pipeline-all} we describe our solution to reducing the number of $k$NN queries during retrieval.
Section~\ref{sec:experiments} evaluates our solution on different feature types and datasets. Section~\ref{sec:conslusion}
concludes this work.

\section{Problem Definition}
\label{sec:problemdef}
Let $DB = \{I_0, ..., I_{|DB|}\}$ denote a database of images $I_j$. Images are represented by a list of interest points and their corresponding
feature vectors, i.e. $I_j = \{p_j^0, ..., p_j^{|I_j|}\}$ with $p_j^i = (v_j^i,x_j^i,y_j^i,s_j^i,r_j^i,\sigma_j^i)$ for affine-variant interest point
descriptors and $p_j^i = (v_j^i,x_j^i,y_j^i,s_j^i,r_j^i,\sigma_j^i,A_j^i)$ for affine-invariant descriptors, with $v_j^i$ a (real-valued
or binary) feature vector, $(x_j^i,y_j^i)$ the coordinate of the interest point in the image, $s_j^i$ its scale, $r_j^i$ its rotation, $\sigma_j^i$
its response, and for affine-invariant descriptors $A_j^i$ the parameters of the ellipse describing its affine shape, see
\cite{PerChuMat09}.

Given a query image $I_q$ containing an object $o$, we would like to retrieve all images $I_n \in DB$ containing object $o$. This is usually achieved
by a combination of \textit{feature matching} and \textit{scoring}. During feature matching, we retrieve tuples of similar feature vectors $m(p_q^i) =
\{(p_q^i,p_{x_0}^{j_0}), \ldots, (p_q^i,p_{x_r}^{j_r})\}$, $\{I_{x_0}, ..., I_{x_r}\} \subseteq DB $ denoting that feature $i$ of the query is
visually similar to the features $\{p_{x_0}^{j_0}, ..., p_{x_r}^{j_r}\}$. This matching problem can for example be solved using the BoVW approach.
In recent years however, as mentioned in the introduction, there has been a shift away from BoVW towards more accurate, however less efficient $k$NN
queries, leading to $m(p_q^i) = \{(p_q^i,p_x^i) | p_x^i \in kNN(p_q^i, DB)\}$ where $kNN(p_q^i, DB)$ retrieves the $k$ tuples from the database whose
feature vectors are closest to the query feature vector given a pre-defined distance function, e.g Euclidean distance. Now let $M = \cup_{i =
0}^{|\Theta|}m(p_q^i)$, with $\Theta = \{p_q^0, \ldots, p_q^{|\Theta|}\} \subseteq I_q$ a subset of the query features. The score of database image
$I_x \in DB$ is computed as $\sum\limits_{\{(p_q^i,p_y^j) \in M | x=y \}}score(p_q^i,p_y^j)$. The most trivial solution would be to increase the score
of image $I_x$ by one for each tentative match, resulting in $\sum\limits_{\{(p_q^i,p_y^j) \in M | x=y \}}1$. More sophisticated scoring approaches
for $k$NN-based image retrieval can be found e.g. in \cite{JegDouSch11b}.

Accurate $k$NN queries are, even after astonishing research efforts in the last years, still relatively expensive. For example, running a 100NN-query
on 100 million binary features using Multi-Index Hashing \cite{NorPunFle14} would take about 100ms, summing up to 100 seconds in a scenario where 1000
features are queried to retrieve a single image\footnote{Note that the number of features extracted from an image is often even larger, see the
dataset statistics in our experimental evaluation}. SIFT features are generally queried approximately, runtimes vary significantly with recall and are
often between 8ms/query and 53 ms/query for a billion SIFT features at a recall below 0.5 \cite{KalAvr14}. Generally, achieving good recall over 0.5
for 1NN queries with such techniques is very expensive. We are not aware of recall evaluations of these techniques for $k\not = 1$ although it was
shown in \cite{JegDouSch11b} that a larger $k$ can notably boost recognition performance. Based on these observations we argue that in addition to
indexing efficiency, other possibilities must be considered to reduce the complexity of the feature matching phase. Generally, to achieve this
complexity reduction, different approaches are reasonable:
\begin{compactitem}
	\item Reduce the \textit{dimensionality} of feature vectors. One well-known approach would be to apply PCA to SIFT features and drop the dimensions
	with least variance. Another more desirable option would be to directly extract lower-dimensional features.
	\item Reduce the \textit{cost} of distance functions, for example by binarization
	\cite{TorFerWei08,JolBui11,ZhoLuLiTia12,HeoLeeHeChaYoo12,HeWenSun13} or by extracting binary features \cite{TrzChrFuaLep13,RubRabKonBra11} and using
	the Hamming distance.
	\item Reduce the \textit{accuracy} of a matching query. This has been widely used in the past, e.g BoVW \cite{SivZis03} can be seen as an extreme
	case.
	\item Reduce the \textit{cost for querying}. A variety of (exact) indexing techniques have been proposed, e.g. Multi-index-hashing \cite{NorPunFle12} for binary features.
	\item Reduce the \textit{number} of $k$NN queries, e.g. \cite{HarHavSch14,HajZha13,LeeKimKimKimYoo10,BroSzeWin05}.
\end{compactitem}

In this paper we focus on the last approach: Let a database of images, represented by sets of features describing the neighborhood around interest points, be given.
Let $n$ denote the upper bound on the number of matching queries, constraining the number of $k$NN queries. The goal of this research is to
develop a retrieval algorithm that returns a list of images ranked by their visual similarity to the query. We aim at modifying the image recognition
pipeline such that a given performance measure (in our case MAP) is maximized for a given $n$.

The problem setting is similar to BoVW-based approaches, however in such a context it is usually assumed that $n=|I_q|$. In this paper we address the
opposite case where $n<<|I_q|$.

\section{Related Work}
\label{sec:related}
This section, addressing related research, follows the organization of the image processing pipeline used in
Section~\ref{sec:pipeline-all}.

\textbf{Keypoint reduction.} In order to reduce the number of extracted features that have to be matched, \cite{HarHavSch14} aimed at predicting the
matchability of features by interpreting the problem as a classification task. Keypoint reduction can also be achieved by employing the Adaptive
Non-Maxima suppression (ANMS) from Brown et al. \cite{BroSzeWin05}.
Their approach aims at finding interest points that are sufficiently distributed across the whole image and is computationally relatively inexpensive.
Hajebi and Zhang \cite{HajZha13} propose to keep track of the distribution of scores during query processing and stop the investigation of further
features as soon as the score difference between the best-scored image and the average score becomes large enough. Other approaches to rank features
are based on visual attention \cite{LeeKimKimKimYoo10}. In contrast to us, the authors query all features of higher scale levels to build a
coarse-grained (32x32) top-down attention map and combine it with a bottom-up saliency map. Then, in an iterative fashion, the features in the most
promising cells of these attention maps are queried. The authors perform some kind of geometric verification, but no match expansion.

\textbf{$k$NN indexing.}
As exact $k$NN query processing on high-dimensional features often cannot significantly decrease runtimes compared to a linear scan due to the curse
of dimensionality, indexing research in the image community concentrates on \textit{approximate} nearest neighbor search. Some well-known approximate
indexing techniques used in image retrieval are forests of randomized $k$D-trees\cite{SilHar08,MujLow09} and the $k$Means-tree
\cite{NisSte06,MujLow09}. These techniques however suffer either from high storage complexity if the database descriptors are needed for refinement,
or low-quality distance approximations. Recent research in $k$NN indexing aims
at providing low runtime and storage complexity while providing accurate distance approximations at the same time. One group of these techniques is
based on the Product Quantization approach from Jegou et al. \cite{JegDouSch11a}, a quantization-based approximate indexing technique distantly
related to the BoVW paradigm. Recent extensions of this approach include \cite{BabLem12,GeHeKeSun13,KalAvr14}. Another group of techniques aiming at
efficient indexing is built on the idea of generating distance-preserving binary codes from real-valued features, sometimes referred to as
\textit{binarization}. Recently developed binarization techniques include the approach
from \cite{TorFerWei08}, Random Maximum Margin Hashing \cite{JolBui11}, Scalar Quantization \cite{ZhoLuLiTia12}, Spherical Hashing
\cite{HeoLeeHeChaYoo12} and k-means hashing \cite{HeWenSun13}. In contrast to binarization techniques, binary keypoint descriptors such as BinBoost
and ORB \cite{TrzChrFuaLep13,RubRabKonBra11} can avoid the indirection of extracting real-valued (e.g.
SIFT) features first and then binarizing them.
Nearest Neighbor queries on databases of binary features can be speeded up by employing (approximate) LSH-based techniques \cite{IndMot98} or exact
indexing \cite{NorPunFle12} and are relatively fast due to them employing the Hamming
distance instead of the Euclidean distance.

\textbf{$k$NN-based Matching.} $k$NN-based matching techniques have a long history in the context of Image retrieval. One of the most famous
techniques using such approaches is Lowe's SIFT recognition pipeline \cite{Low04}. Lowe retrieved, for each query feature, the two nearest neighbors
from the database and accepted a feature as match if its distance ratio between 1NN and 2NN was above a given threshold.
J\'{e}gou et al. \cite{JegDouSch11b} evaluated $k$NN-based matching based on local features, especially SIFT. They proposed a voting scheme optimized
for $k$NN-based retrieval. This adaptive criterion basically scores matches relative to the distance of the $k$-th match. Furthermore, the authors
analyzed normalization methods for the resulting votes in order to reduce the negative effect of favouring images with many features over those with
only a few. They did however not consider reducing the number of query features. Qin et al. \cite{QinWenGoo13} proposed a normalization scheme for
SIFT-features that locally reweights their Euclidean distance, optimizing the separability of matching and non-matching features. Based on this
normalization, the authors developed a new similarity function and scoring scheme based on thresholding rather than $k$NN query processing.

\textbf{Match Expansion.}
As our technique aims at reducing the number of $k$NN queries during the matching step, the generation of a sufficient number of match hypotheses has
to be achieved in a different fashion. We do so by applying a flood-filling approach using $k$NN matches as seed points.
Match expansion has received quite some attention in the computer vision community \cite{SchMoh96,SchZis02,SivZis03,FerTuyGoo04,GuoCao12,CuiNga13},
and will most likely become more relevant again with the use of $k$NN-based matching techniques. One of the first technique in this area of research
has been proposed by Schmid and Mohr \cite{SchMoh96}.
They used the spatial neighbors of match candidates to increase the distinctiveness of features. They also considered the consistency of gradient angles between these
features to reject false-positive matches, however they did not consider the combination of their approach with feature reduction. Sivic and Zisserman
adapted the technique for Video Google \cite{SivZis03}. We however do not reject matches based on this technique but rather increase the score of a
given image by considering neighboring features.
Our work is also inspired by \cite{SchZis02}, where the authors used a region-growing approach for establishing correspondences in the context of
multi-view matching. After establishing a set of initial matches in a traditional index-supported manner, an affine transformation is estimated that
guides search of additional matches in a local neighborhood of the seed match. The authors, however, did not use this technique for reducing the
number of queries in the matching step, but rather to increase the result quality.
Ferrari et al. \cite{FerTuyGoo04} developed another related technique in order to achieve high invariance to perspective distortion and non-rigid
transformation; it further allowed to perform an accurate segmentation of objects during recognition. Their approach builds a dense grid of features
over the image; in contrast we use the initially provided keypoints and descriptors that are stored in the database nonetheless, reducing
computational overhead. A recent work related to this approach includes \cite{CuiNga13}. Guo and Cao \cite{GuoCao12} proposed to use Delaunay
triangulation to improve geometric verification.
Wu et al. \cite{WuKeIsaSun09} proposed to enrich visual words by their surrounding visual words, generating scores not only by the weight of a visual
word, but also the neighboring features; the authors however did not consider keypoint reduction. 
Geometric min-Hashing \cite{ChuPerMat09}, based on the BoVW-paradigm, considers neighboring features as well, however in the context of
hashing. The approach aims at increasing precision at the cost
of recall, by dropping features that do not share a similar neighborhood. However, if we reduce the number of matching queries, one of the main
concerns is recall, such that our approach aims at increasing MAP without negatively affecting recall.

\section{Pipeline}
\label{sec:pipeline-all}
The general retrieval pipeline from this paper follows the one used in the past for BoVW-based image retrieval, but in order to
incorporate $k$NN queries and reduce the number of query features we applied some changes. In this section, we first provide a theoretic overview over the pipeline. Then, as implementing the pipeline in such a naive way would lead to unacceptable overhead in terms of memory and computational
resources, we provide practical considerations about its implementation in a real-world setup.

\subsection{Theory}
\label{sec:pipeline}
We split our pipeline into the stages of feature detection and extraction, feature ranking, feature matching, match expansion, scoring, and
re-ranking. The pipeline was designed with extensibility in mind such that each stage, e.g. keypoint reduction and match expansion, can be easily
exchanged by different techniques. 

\paragraph{1) Feature Extraction.}During feature extraction, given the query
image, we extract the set $I_q$ of keypoints and descriptors. Possible features include floating point features such as SIFT \cite{Low04} or binary features such as BinBoost and ORB \cite{TrzChrFuaLep13,RubRabKonBra11}.
The cardinality of $I_q$ depends on the used feature extractors and can range up to several thousand features.

\paragraph{2) Feature Ranking.}The next stage, feature ranking, is based on the
idea that some features in an image contain more information than others. For example, vegetation usually provides less information about a specific object contained in the image than the features of the object
itself. We aim at ordering the extracted features by a given quality measure, as we would like to query the most promising features first, i.e. the features
with the highest chance of providing good match hypotheses. There exist several techniques for feature ranking, and we will fall back to these instead
of developing a new approach. The only criterion such a technique needs to fulfill in order to be integrated into the recognition pipeline is that it
returns a quality score for each query feature. A simple baseline is a random ranking. Features can also be ranked by their response or size.
More sophisticated techniques include Adaptive Non-Maximal Suppression \cite{BroSzeWin05} and the use of decision trees involving additional training
\cite{HarHavSch14}, which has however neither been adapted to binary features nor to $k$NN-based matching, yet. The result of this feature ranking
step is a feature list, ordered such that the most promising features appear first.

\paragraph{3) Feature Matching.}The next step, feature matching, aims at finding
match hypotheses for the highest ranked features found during the last step. For each of the first $n$ features in the ranking, a $k$NN query is posed on the database. 
The selection of the parameter $k$ of the $k$NN query is important for maximizing the quality of the query result \cite{JegDouSch11b}. On the one hand
side, a large $k$ decreases the quality of the query result, as this introduces a high number of erroneous correspondences which have to be filtered
out during a verification step later in the pipeline. On the other hand, a small $k$ also reduces the retrieval quality as many high-quality
hypotheses are left unconsidered. Basically, $k$ can be seen as a way to tweak \textit{recall} at a given number of query features, as the number of
images returned by the query is at most $n*k$. As a result, especially if a
very small number of $k$NN queries is used for correspondence generation, it is possible that an even larger $k$ increases effectiveness, as it allows for finding more initial correspondences
(however of lower quality). We refer to Section~\ref{sec:experiments} for an experimental analysis of this problem. The feature matching stage
provides a list of tentative matches (tuples) $(p_q^i,p_{x}^j)$.

\paragraph{4) Match expansion.}The match expansion phase is tightly interleaved
with the match generation phase. In our scenario where we want to pose a very small number of $k$NN queries on the system, we face the problem that even if we find some correspondences between the query and a database
image, their number will be relatively low, increasing the probability that a good match is outranked by an image containing common random matches
only. To resolve this problem, we shift the load of correspondence generation from the matching stage --that employs $k$NN queries-- to an
intermediate stage that avoids such queries. 
Match expansion aims at reducing the runtimes of generating additional matches, which usually depend on the underlying index structure, to runtimes
depending on the features stored in a single image pair. When employing exhaustive search with product quantization for indexing, match expansion
therefore avoids additional linear scans over the feature database; as non-exhaustive variants of product quantization only consider a fraction of
features in the database, the gain of match expansion in this case depends on the desired recall of the index structure.

It is however important to realize that, while such a match expansion can
find additional hypotheses for candidate images, i.e. increase \textit{MAP}, it cannot retrieve any new candidates, i.e. increase recall. This
stage therefore aims at compensating for the loss in MAP due to querying less features.

\begin{figure*}[t] \centering
\includegraphics[width=\textwidth]{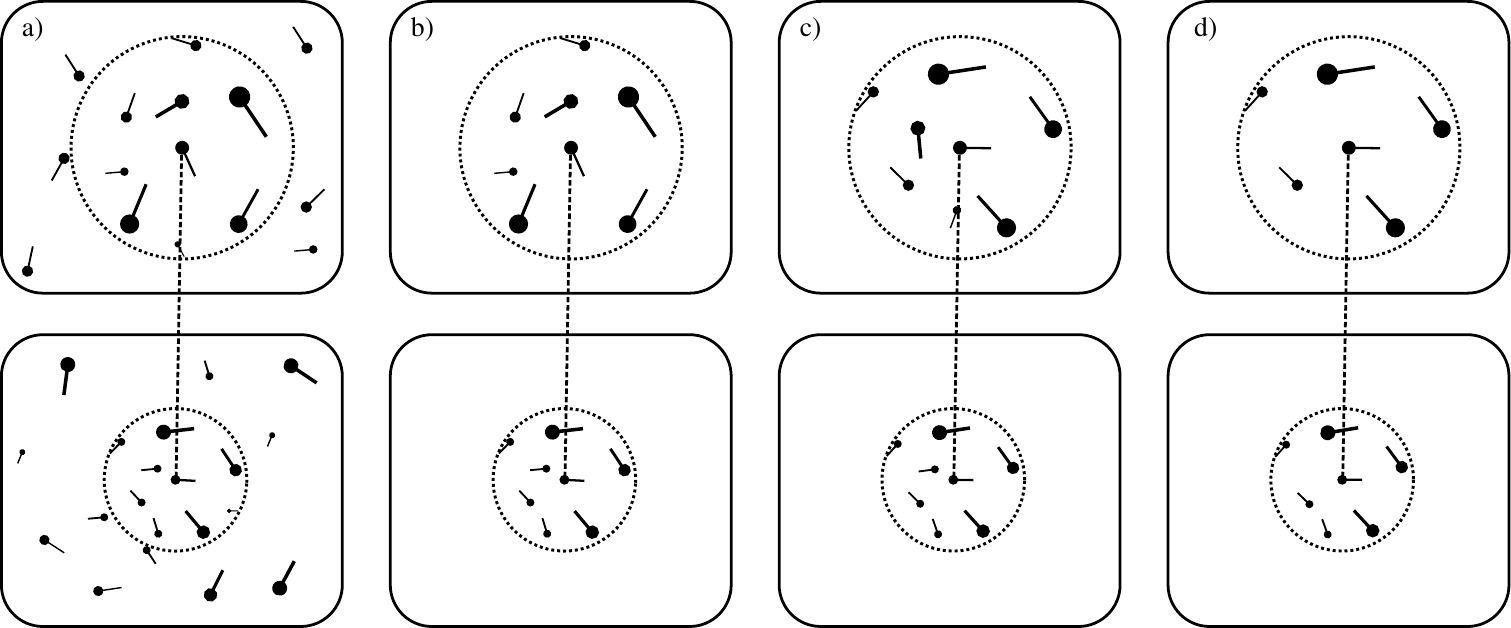}
  \caption{Generation of additional match hypotheses.}
  \label{fig:hypothesis_generation}
\end{figure*}

Match expansion exploits the keypoint information of the seed matches that
provide scale, rotation, and possibly affine information.
These properties can be used to identify spatially close keypoints, adapting the
ideas of \cite{SchZis02,FerTuyGoo04,ChuPerMat09,WuKeIsaSun09}; we will use a
modified version of \cite{SchMoh96} for expanding matches.
Given that a match hypothesis is correct, not only the corresponding feature
pair should match, but also its spatial neighborhood, as an object is usually
not only described by a single but rather by multiple keypoints. The similarity
of a match's neighborhood is evaluated using the procedure visualized in
Figure~\ref{fig:hypothesis_generation}. The figure shows an initial seed match,
i.e. a $k$NN of a query feature, and keypoints surrounding the seed match. The
scale of each keypoint is represented by the keypoint's size, and the gradient
direction is represented by a line anchored in the keypoint's center. The top
row of this figure visualizes the features of the query image, while the bottom
row visualizes the image features of a tentative match.

Starting point is an initial correspondence pair $(p_q^i, p_{\db}^j)$ established by $k$NN-search in feature space, see
Figure~\ref{fig:hypothesis_generation}~a). In a first step, features in a given spatial range are retrieved in the image $I_q$ for $p_q^i$ and in
Image $I_{\db}$ for $p_{\db}^j$, see Figure~\ref{fig:hypothesis_generation}~b); the spatial range is visualized by a dotted circle. Given the constant $\delta_{xy}$, the spatial range is given by $s_q^i\delta_{xy}$ for the query feature
and $s_{\db}^j\delta_{xy}$ for the matching database feature, achieving scale invariance. Spatially close keypoints with a significantly different
scale (determined by the scale ratio threshold $\delta_{s}$) than their reference feature are discarded (see the small features in the figure) similar to
\cite{ChuPerMat09}, resulting in two sets of features $P_q$ and $P_{\db}$. These remaining features are rotation-normalized using the reference
keypoint's gradient orientation information $r_q^i$ and $r_{\db}^i$, rotating the set of keypoints and their corresponding gradient orientations, see
Figure~\ref{fig:hypothesis_generation}~c).
Then the two lists of keypoints are traversed in parallel. If the rotation-normalized angle $\alpha$ to the reference feature, the rotation-normalized
gradient angle $r$, and the feature-space distance of two features $d_v$ are within a predefined threshold ($\delta_\alpha$, $\delta_r$, and
$\delta_{d_v}$ respectively) and the ratio of their scale-normalized spatial distance is within given bounds $\delta_{d_{xy}}$, the corresponding
features are accepted as a matching pair (see Figure~\ref{fig:hypothesis_generation}~d)). The remaining features are discarded. Note
that, while the complexity of this step is $|P_q|*|P_{\db}|$ in the worst case, it can be reduced by an efficient sweep-line implementation that sorts features by
their angle $\alpha$ and traverses both lists in parallel.

This technique of finding neighboring keypoints assumes that two images are only distorted by similarity transforms. To mitigate the effects of
non-similarity or even (small) non-rigid distortions, a recursive procedure (in our case with a maximum recursion depth of~2) can be chosen that
performs the same procedure on each of the resulting pairs. Moreover, by choosing the Mahalanobis distance using the affinity matrices of the seed
pair ($A_q^i$ and $A_{\db}^i$ respectively) instead of Euclidean distances for finding spatially neighboring keypoints, the process can be extended to
affine-invariant features. This technique returns features within an elliptical region around the seed points, reducing performance loss from affine
distortions.

Result of the expansion phase is an extended list of match hypotheses.

\paragraph{5) Scoring.}Scoring is again tightly interleaved with match
generation.
In this phase, based on the expanded list of matches, a score is computed for every database image. In the simplest case, each hypothesis pair votes with a score of one for a given database image. This however, has
shown to have a relatively low performance \cite{JegDouSch11b}, as for example images containing many features would have higher scores than images
containing only a few features. For this purpose, more sophisticated scoring techniques have been developed. We will adapt some of the techniques from
\cite{JegDouSch11b}, weighting scores based on the distance of the candidate
feature to the query feature and the number of features in the image. For each
matched feature from image $I_x$ its score is increased by $\frac{\sqrt{d_{kNN}-d_{ref}}}{\sqrt{|I_q|}\sqrt{|I_x|}}$ with $d_{kNN}$ the $k$NN distance of the seed feature, and $d_{ref}$ the distance between the seed feature and its tentative match in the candidate image, i.e. features generated during match
expansion are assigned the same score as their seed match. This score is similar to the scores from \cite{JegDouSch11b}, however we have added
additional square root weighting which further increased effectiveness of these scores. For scoring we implemented a simple burst removal
\cite{JegDouSch09} scheme after match expansion that allows only for one correspondence per query feature.

\paragraph{6) Re-Ranking.}After building match hypotheses and scoring, the
ranked list can be processed equivalently to BoVW-based approaches. Further
steps can include geometric verification or query expansion techniques
\cite{ChuPhiSivIsaZis07}. As these techniques are complementary to the remaining
pipeline we will not further consider them in this chapter.

\subsection{Practical Considerations}
To enable efficient query processing using the pipeline summarized previously, three conditions must be fulfilled. First, it must be possible to
efficiently retrieve the $k$NN features of a query feature and their corresponding keypoints from the database. Second, to enable match expansion, it
must be possible to compute, given two keypoints, the distance of their corresponding feature vectors. Third, also concerning match expansion, it must
be possible to pose a range query on all keypoints from a given image, retrieving spatially close keypoints. In the most basic case, the image
database used for query processing can be seen of a list of tuples $(p_0^0, \ldots, p_0^{|I_0|}, \ldots, p_i^0, \ldots, p_i^{|I_i|}, \ldots)$
containing feature and keypoint information. The features in the list are ordered by their corresponding image to allow efficient match expansion.
However, in order to enable usability of this approach in a practical setup, special care has to be taken concerning computational and memory
efficiency and the thorough selection of parameters; we will address solutions for these challenges in the following section. Computational efficiency
can be achieved using indexing techniques such as Product Quantization or Multi-Index Hashing, while the memory footprint of the image database can be
reduced by compressing the feature vectors used during match expansion. Finally, the selection of parameters can be achieved using appropriate
optimization techniques.

\subsubsection{Indexing}
In order to improve the performance of of the pipeline in real-world
applications, fast (approximate) indexing techniques optimized for
high-dimensional data
\cite{KalAvr14,GeHeKeSun13,BabLem12,NorPunFle12,JegDouSch11a} can be employed.
In this research we focused on (Locally Optimized) Product Quantization for
real-valued features and Multi-Index Hashing for binary features; we will
summarize these techniques in the following paragraphs for the sake of
completeness.

\paragraph{Product Quantization.} Approximate nearest neighbor search based on
Product Quantization, initially proposed by J\'{e}gou et al.
\cite{JegDouSch11a} and further optimized e.g.
in \cite{KalAvr14,GeHeKeSun13,BabLem12}, is an elegant solution for indexing
high-dimensional real-valued features. During a training phase, features in the
database are clustered using $k$-means and the database features are assigned to
their closest cluster mean, partitioning the set of vectors into distinct cells,
similar to Locality-Sensitive Hashing \cite{IndMot98}. Then, for each feature
vector, the residual to its corresponding cluster mean is computed and the
resulting residuals are product quantized. Product quantization is achieved by
splitting a vector into a small number of subvectors (e.g. 8) and quantizing
each of these subvectors separately using a relatively small codebook of e.g.
256 centroids. Instead of storing the residuals themselves, only the cluster id
of the closest residual is stored in the index for each subvector, resulting in
a reduction in memory complexity. With product quantization using 8 subvectors
of 256 cluster centers, a SIFT vector could be compressed from 128 bytes to 8
bytes, resulting in a compression of nearly 95\%.
The index itself consists mostly of a list of outer clusters and for each of
these clusters an inverted list storing, for each feature assigned to this
cluster mean, its list of quantized subvectors.

During query evaluation, the query is first assigned to the closest outer
cluster mean (or possibly the closest $c$ means in the case of
multi-assignment). Then the inverted lists of these means are scanned, and a distance approximation is computed for each of the database vectors stored in this list: As vectors are represented as a list of their closest subvector-centroids, a distance approximation can be generated by summing the squared distances
of the corresponding centroids which can be sped up with the use of look-up
tables. The resulting distance approximations are then used to rank the feature vectors.
In the past, a variety of improvements of this approach have been proposed, for example the Inverted Multi-Index \cite{BabLem12}, Optimized Product
Quantization \cite{GeHeKeSun13}, and Locally Optimized Product Quantization (LOPQ) \cite{KalAvr14}. For our experiments we will use the most recent of
these approaches, namely LOPQ.

\paragraph{Multi-Index-Hashing.} While Product Quantization has been developed to support efficient query processing on real-valued and high-dimensional
feature vectors such as SIFT, Multi-Index Hashing (MIH) \cite{NorPunFle12} has been specifically designed for binary features, such as ORB or
BinBoost\cite{TrzChrFuaLep13,RubRabKonBra11}. It is based on the idea of Locality-Sensitive Hashing \cite{IndMot98}, however in contrast to this
approach it aims at exact query processing. The idea behind MIH is, similar to Product Quantization, to split a binary vector into a set of
subvectors. Each of these subvectors is indexed in a dedicated hash table with the subvectors' binary value directly representing the id of its hash
cell: A single cell of the index contains all database vectors that contain a given subvector.

During query processing, the query is split into subvectors as well. These
subvectors provide the hash cells that have to be looked up in order to find
vectors with similar values. Bit-flipping the query subvectors and retrieving
the corresponding hash cells allows retrieving features with similar, but not
equivalent subvectors. To allow exact $k$NN processing, Norouzi et al. developed
a retrieval strategy that enumerates all relevant bit-flip operations to
retrieve an exact query result. In our experimental evaluation, we will use this
index structure in combination with BinBoost\cite{TrzChrFuaLep13} features to
evaluate the pipeline from Section~\ref{sec:pipeline} on binary
features.

\subsubsection{Match Expansion}
Concerning the expansion of initial matches we face two challenges. First, we have to find the best parameters for the expansion step. Second, memory
consumption has to be minimized in order to store features in main memory and hence speed up query processing.

\textbf{Parameter Selection.} 
Unfortunately it is a tedious task to determine the thresholds of the flood-filling procedure for match expansion, namely $\delta_{d_v}$,
$\delta_\alpha$, $\delta_r$, and $\delta_{d_{xy}}$, by hand. This problem can be solved by utilizing Nelder-Mead Simplex-Downhill optimization:
After selecting the distance multiplier $\delta_{xy}$ and the maximum scale change ratio $\delta_{s}$ by considering runtime constraints, the
remaining thresholds are automatically determined by the Simplex-Downhill approach. Optimization of these parameters should be conducted on a training
dataset different from the test set in order to avoid overfitting.

\textbf{Vector Compression.} 
For compressing real-valued feature vectors, we consider Product Quantization as
well. In contrast to Product Quantization based indexing based on LOPQ, however,
we do not product quantize residual vectors, but rather the vectors themselves, as otherwise vectors belonging to different cells in the outer quantizer could not be compared efficiently. For compression, we split each feature vector in a set of $m=8$ subquantizers and for each of these subquantizers
build a codebook of $s=256$ centroids. The distance between feature vectors can then easily be approximated as the sum of squared distances between
the closest subquantizer centroids followed by a square root operation. As distances between cluster centroids can be stored in a lookup table
of size $m*s*s$, distance computations reduce to $m$ table look-ups and a single
square root operation.
\section{Experiments}
\label{sec:experiments}

\subsection{Experimental Setup}
\textbf{Datasets. }We evaluated the modified recognition pipeline on four
datasets. The \textit{Oxford5k} (O5k) building dataset \cite{PhiChuIsaSivZis07} consists of 5063 images of
common tourist landmarks in Oxford. The authors of the benchmark also provide a
set of 55 queries including rectangular query regions and ground truth files
listing, for each query, the images that contain at least parts of the query.
Ground truth files are split into three categories: good, ok and junk. Good and
ok files are considered for computing the Mean Average Precision (MAP) of the
query. Junk images are neither scored as true hit nor as false hit and simply
discarded for computing the MAP. We also included \textit{Oxford105k} (O105k) in
our evaluation which consists of the Oxford5k dataset in combination with about
100k distractor images \cite{PhiChuIsaSivZis07} that do not contain images
related to the query.
The \textit{Paris6k} (P6k) dataset \cite{PhiChuIsaSivZis08}, conceptually
similar to the Oxford dataset, consists of 6412 images of common landmarks in
Paris, and has the same structure as the Oxford dataset. As a third dataset we
used the \textit{INRIA Holidays} (Hol) dataset \cite{JegDouSch08} which consists
of 1491 images including 500 queries and their corresponding ground truth. In
contrast to the Oxford and Paris dataset, Holidays contains more natural scenes
and a lower number of result images for each query. Images of the Holidays
dataset were scaled down to a maximum side length of 1024 before feature
extraction. 

\textbf{Feature Extraction and Indexing. }We used two different feature extraction techniques: a rotation-variant version of SIFT using affine
invariant keypoints\footnote{https://github.com/perdoch/hesaff/} made available
by the authors of \cite{PerChuMat09} and, as an instance of state-of-the-art
binary descriptors, the BinBoost descriptor which is also publicly available \cite{TrzChrFuaLep13}. We decided to include binary features in our
evaluation as we see them as another mean of decreasing query complexity, however we will concentrate on SIFT features in our evaluation.

\begin{table}
   \center
   \caption{Database Statistics}
   \label{tab:data_statistics}
   \begin{tabular}[h!]{|c||c|c|c|c|}
      \hline
      Dataset & Extractor & Features & $\diameter$ \\
      \hline
      \hline
      O5k   & BinBoost          & 10,640,081   & 2101.5\\
      O105k & BinBoost          & 195,068,373  & 1855.4\\
      O5k   & SIFT (Hess.-Aff.) & 13,516,675   & 2669.7\\
      O105k & SIFT (Hess.-Aff.) & 253,761,866  & 2413.7 \\ 
      P6k &  SIFT (Hess.-Aff.) & 16,073,531 & 2506.8\\
      Hol & SIFT (Hess.-Aff.) & 4,596,567 & 3082.9 \\ 
      \hline
   \end{tabular}
   \vspace{-1em}
\end{table}

Concerning \textit{Hessian-affine SIFT}, scale was separated from the affinity
matrices according to \cite{PerChuMat09}, however for expanding matches we used
the square root of this scale which roughly corresponds to the radius of the
image patch used for SIFT extraction.
The parameters of the feature extraction stage have been left at the default
parameters. SIFT features are 128-dimensional real-valued vectors.
These vectors were square-root weighted similar to RootSift\cite{AraZis12},
however without $l_1$ normalization. The weighted features were then indexed using LOPQ in combination
with a multi-index \cite{KalAvr14}. We use a vocabulary of size $V=2*1024$ for
the inverted lists, and 8 subquantizers for vector quantization, each
subquantizer with a vocabulary of size of 256 clusters.
The corresponding source code has been kindly provided by the
authors\footnote{http://image.ntua.gr/iva/research/lopq/}. To compress the
feature vectors for the expansion phase, we again used 8 subquantizers
consisting of 256 clusters, reducing storage overhead of feature vectors to
6.25\% of their uncompressed memory footprint. Codebooks for the Oxford and
Holidays datasets were trained on Paris6k, and for Paris6k code books where
trained on Oxford5k. During query processing, we applied a simple means of
burst removal \cite{JegDouSch09}, scoring each query feature once even if it had more
than one match.

\begin{table*}
   \center
   \caption{Parameters for Match Expansion}
   \begin{tabular}[h!]{|c||c|c||c|c|c|c|c|} 
      \hline
      Extractor &  Train   & $\delta_{xy}$ & $\delta_s$ & $\delta_{d_v}$ & $\delta_\alpha$ & $\delta_r$ & $\delta_{d_{xy}}$\\
      \hline
      \hline
      SIFT & P6k & 6              &     0.8    &      26.2    &     24.3       &   --          & 0.49\\
      \hline
      SIFT & O5k & 6              &     0.8    &      26.9    &     18.9       &   --          & 0.56\\
      \hline
      BinBoost       & P6k  & 4             &     0.8    &     73        &    21.1       &   26.0        & 0.46 \\   
      \hline
   \end{tabular}
   \label{tab:parameters}
\end{table*}

\textit{BinBoost} descriptors, i.e. 256-dimensional binary vectors, can be queried rather efficiently using exact indexing techniques optimized for
binary feature vectors, e.g. \cite{NorPunFle12}. We have used a publicly available implementation of their index during our experimental evaluation.
We applied burst removal when querying these features as well.

An overview of the extracted features can be found in Table~\ref{tab:data_statistics}. Note that the number of query features was different to the
number of database features on Oxford5k, Oxford105k and Paris6k due to the bounding boxes provided by the dataset authors, and for several queries the
number of query features was less than 1000. The average number of features available over all queries was 1371.4 (BinBoost,
$\sigma=612.3$) and 1452.8 (SIFT Hessian-Affine, $\sigma=950.2$) for queries on Oxford.

The code was written in C++ using OpenCV. Runtime experiments were conducted on
an off-the-shelf Linux Machine with i7-3770@3.40GHz CPU and 32GB of main memory without parallelization. During our experimental evaluation we
concentrate on analyzing the effectiveness of the approaches in terms of Mean Average Precision (MAP); we also provide numbers on the performance of the evaluated approaches concerning the runtime of the scoring, querying
and ranking stages.

\textbf{Parameters.} The parameters for query processing were set as follows. First, range multiplier $\delta_{xy}$, maximum
scale change $\delta_s$, $k$, and $n$ were set by hand with computational efficiency in mind, as a lower number of features considered during
expansion reduces the cost of this step. Given these manually set parameters, the remaining parameters of the expansion phase, i.e.
feature distance threshold $\delta_{d_v}$, angular threshold $\delta_\alpha$, gradient angle threshold $\delta_r$ and spatial distance ratio $\delta_{d_{xy}}$
were set to the outcome of a Nelder-Mead Downhill-Simplex optimization maximizing MAP; initialization was performed with reasonable seed values.
Minimization was done on the Paris6k dataset (with LOPQ and quantization code books trained on Paris6k as well) for the Oxford5k, Oxford105k and Holidays datasets. For the Paris6k dataset, we optimized these parameters on the Oxford5k dataset. The parameters were selected for each of the
descriptor types (SIFT and BinBoost) using ANMS ranking at $k=100$, number of keypoints $n=10$, recursively descending into every expanded match. The resulting parameters
were reused for the remaining ranking approaches, different $k$, $n$ and the non-recursive approach. An overview over the selected parameters
is shown in Table~\ref{tab:parameters}.

We varied each of the optimized parameters by $\pm 10\%$ separately on Oxford 5k (ANMS ranking with match expansion) to get insights into their
effect on MAP. The maximum deviation resulted from decreasing the feature distance threshold, which lead to a decrease in MAP of $-0.012$, indicating
that while there is an impact of the optimized parameters on the performance of match expansion, there is still a range of relatively ``good''
parameters.

\subsection{Experiments}
We evaluated the algorithm's performance by varying $k$ and $n$ as these parameters affect the number of initial seed points that are expanded later.
As a baseline for our experiments we implemented a scoring scheme based on \cite{JegDouSch11b} that considers the distances between features and the
number of features in the image for score computation.

\begin{table}
   \center
   \caption{SIFT, Oxford5k, k=100}
   \label{tab:k100SIFT_HESAFF_OXFORD5K_SC6_LOPQ_RSNL1_SQ8}
   \begin{tabular}[h!]{|c||c|c||c|c|}
      \hline
      $\downarrow$ Appr. $\rightarrow n$& 50 & 100 & 500 & 1000 \\
      \hline
      \hline
   RND & .616 & .698 & .810 & .827 \\
   \hline
   RESP & .557 & .640 & .787 & .822 \\
   \hline
   ANMS & .676 & .727 & .825 & .836 \\
   \hline
   \hline
   RND+ME & .679 & .749 & .829 & .838 \\
   \hline
   ANMS+ME & .741 & .780 & .843 & .844 \\
   \hline
   \hline
   RND+MER & .686 & .752 & .826 & .832 \\
   \hline
   ANMS+MER & .752 & .786 & .837 & .838 \\
   \hline
   \end{tabular}
\end{table}

\begin{table}
   \center
   \caption{SIFT, Paris6k, k=100}
   \label{tab:k100SIFT_HESAFF_PARIS6K_SC6_LOPQ_RSNL1_SQ8}
   \begin{tabular}[h!]{|c||c|c||c|c|}
      \hline
      $\downarrow$ Appr. $\rightarrow n$& 50 & 100 & 500 & 1000 \\
      \hline
      \hline
   RND & .566 & .652 & .770 & .786 \\
   \hline
   RESP & .519 & .594 & .743 & .775 \\
   \hline
   ANMS & .578 & .668 & .783 & .794 \\
   \hline
   \hline
   RND+ME & .629 & .699 & .781 & .789 \\
   \hline
   ANMS+ME & .648 & .723 & .793 & .796 \\
   \hline
   \end{tabular}
\end{table}

\textbf{Keypoint Ranking.} In our first experiment (see
Table~\ref{tab:k100SIFT_HESAFF_OXFORD5K_SC6_LOPQ_RSNL1_SQ8} and
Table~\ref{tab:k100BIN_BOOST_OXFORD5K_SC4}) we wanted to evaluate the
performance difference in MAP when querying a low number of features (i.e. 50,
100, 500 and 1000 keypoints) with different keypoint ranking techniques,
providing a baseline for further experiments.
The simplest ranking (RND) takes random features from the extracted keypoints;
we averaged this approach over 5 runs to get accurate results. Furthermore we
evaluated a ranking based on keypoint responses (RESP), and a more sophisticated
approach called Adaptive Non-Maximal Suppression \cite{BroSzeWin05} (ANMS) that
aims at distributing keypoints relatively uniformly over the image. As expected,
considering only few keypoints significantly reduces the MAP of all approaches.
The MAP of the response-based ranking is worse or similar to the random
baseline: for SIFT, the response decreases performance compared to the random
approach, while for BinBoost (that is based on SURF Keypoints) results are
sometimes slightly better than the random baseline. The ANMS ranking increases
the MAP for all approaches. 
Note that the gain resulting from using ANMS is rather astonishing for the
Oxford5k dataset; we can easily gain 0.03 ($n$=100) to 0.06 ($n$=50) points in
MAP without significant computational overhead if the number of features queried is
relatively low. Similar observations hold for Holidays
(Table~\ref{tab:k10SIFT_HESAFF_HOLIDAYS_SC6_LOPQ_RSNL1_SQ8}) but considering
Paris6k (Table~\ref{tab:k100SIFT_HESAFF_PARIS6K_SC6_LOPQ_RSNL1_SQ8}), the gain
resulting from using ANMS instead of a random ranking is lower. Our results with
BinBoost (Table~\ref{tab:k100BIN_BOOST_OXFORD5K_SC4}) on Oxford5k indicate that
ANMS without match expansion can increase performance by over 0.07
points in MAP ($n$=50), however its performance is generally lower than SIFT,
even if SIFT vectors are quantized as in our case; the memory overhead (8 bytes) for
quantized SIFT vectors is actually lower than for BinBoost (32 bytes) features.

\begin{table}
   \center
   \caption{SIFT, Holidays, k=10}
   \label{tab:k10SIFT_HESAFF_HOLIDAYS_SC6_LOPQ_RSNL1_SQ8}
   \begin{tabular}[h!]{|c||c|c||c|c|}
      \hline
      $\downarrow$ Appr. $\rightarrow n$& 50 & 100 & 500 & 1000 \\
      \hline
      \hline
   RND & .600 & .662 & .765 & .792 \\
   \hline
   RESP & .571 & .630 & .735 & .770 \\
   \hline
   ANMS & .642 & .696 & .779 & .803 \\
   \hline
   \hline
   RND+ME & .646 & .702 & .764 & .770 \\
   \hline
   ANMS+ME & .699 & .734 & .780 & .781 \\
   \hline
   \end{tabular}
\end{table}

\begin{table}
   \center
   \caption{BinBoost, Oxford5k, k=100}
   \label{tab:k100BIN_BOOST_OXFORD5K_SC4}
   \begin{tabular}[h!]{|c||c|c||c|c|}
      \hline
      $\downarrow$ Appr. $\rightarrow n$& 50 & 100 & 500 & 1000 \\
            \hline
      \hline
	   RND & .390 & .462 & .586 & .616 \\
	   \hline
	   RESP & .389 & .461 & .600 & .625 \\
	   \hline
	   ANMS & .461 & .508 & .614 & .620 \\
	   \hline
	   \hline
	   RND+ME & .469 & .529 & .625 & .638 \\
	   \hline
	   ANMS+ME & .542 & .588 & .648 & .644 \\
	   \hline
	   \hline
	   RND+MER & .481 & .539 & .626 & .634 \\
	   \hline
	   ANMS+MER & .551 & .591 & .648 & .640 \\
	   \hline
	\end{tabular}
\end{table}

\textbf{Match expansion.} Our second experiment aims at evaluating the gain in MAP that can be achieved for a low number of $k$NN queries when
additional hypotheses are generated by match expansion (ME) and the same approach in its recursive version (MER). Affine-invariant SIFT
(ANMS+ME, $n$=50) achieves about 90\% of the random baseline (RND, $n$=1000) at
50 keypoints on Oxford5k, where the baseline only achieves 75\%. At the same time the results at 1000 keypoints are similar for all approaches, showing that match expansion does not considerably affect MAP if a high number of keypoints is
queried. This substantiates our statement made in the introduction: if a small number of features is queried, techniques that do
 not achieve significant performance gain for a high number of features can achieve considerable gain in performance. Results are similar for
Holidays (88\% for ANMS+ME@$n=50$ vs. 76\% for the random Baseline) while for Paris6k the gain of match expansion is lower (82\% vs 72\% for the
random baseline). Further note that MAP for ANMS+ME decreases slower with
decreasing $n$ than without expansion (-0.003 (ANMS+ME) vs. -0.011 (ANMS) for
$n:1000\rightarrow500$ on Paris6k). Results for Oxford105k are shown in Table~\ref{tab:k100SIFT_HESAFF_OXFORD105K_SC6_LOPQ_RSNL1_SQ8}.

Using BinBoost (Table~\ref{tab:k100BIN_BOOST_OXFORD5K_SC4} and Table~\ref{tab:k100BIN_BOOST_OXFORD105K_SC4}) the results are similar. The combination
of ANMS ranking and match expansion at 100 keypoints performs similar to the
random baseline at 500 keypoints, which is especially interesting as the exact indexing techniques used for BinBoost already lead to a relatively high runtime.

While there is some gain for recursively descending (MER) into matches,
this additional step does not significantly improve the performance with both SIFT and BinBoost, while being computationally much more expensive.
Therefore we will concentrate on ME in the following. 
ANMS+ME on Oxford using SIFT accepted about 16500 matches per query image ($n=100$, $k=100$), in
contrast to the approximately 8800 tentative correspondences (less than $n*k$ due to burst removal) that have been generated using $k$NN matching
alone (ANMS).

Note that we also evaluated the effect of a re-ranking stage, weak geometric
consistency (WGC) \cite{JegDouSch08} with 32 angular and 16 scale bins, on match
expansion using BinBoost. Both pipelines, with and without match expansion were
positively affected by WGC, gaining about 0.04 (ANMS and ANMS+ME) points in MAP
at $n=1000$, indicating that WGC is complementary to match expansion. We further
observed that WGC did not have a large effect with a low number of
features (0.013 with ANMS and 0.006 with ANMS+ME at $n=100$)
involved both with and without match expansion.

\begin{table}
   \center
   \vspace{-1em}
   \caption{SIFT, Oxford 105k, k=100}
   \label{tab:k100SIFT_HESAFF_OXFORD105K_SC6_LOPQ_RSNL1_SQ8}
   \begin{tabular}[h!]{|c||c|c||c|c|}
      \hline
      $\downarrow$ Appr. $\rightarrow n$& 50 & 100 & 500 & 1000 \\
      \hline
      \hline
   ANMS & .489 & .554 & .710 & .748 \\
   \hline
   ANMS+ME & .584 & .630 & .753 & .775 \\
   \hline
   \end{tabular}
   \vspace{-1em}
\end{table}

\begin{table}
   \center
   \caption{BinBoost, Oxford 105k, k=100}
   \label{tab:k100BIN_BOOST_OXFORD105K_SC4}
   \begin{tabular}[h!]{|c||c|c||c|c|}
      \hline
      $\downarrow$ Appr. $\rightarrow n$& 50 & 100 & 500 & 1000 \\
      \hline
       \hline
	   ANMS & .369 & .412 & .527 & .558 \\
	   \hline
	   ANMS+ME & .445 & .477 & .576 & .590 \\
   \hline
   \end{tabular}
\end{table}

\begin{figure}[t] \centering
  \includegraphics[width=\columnwidth]{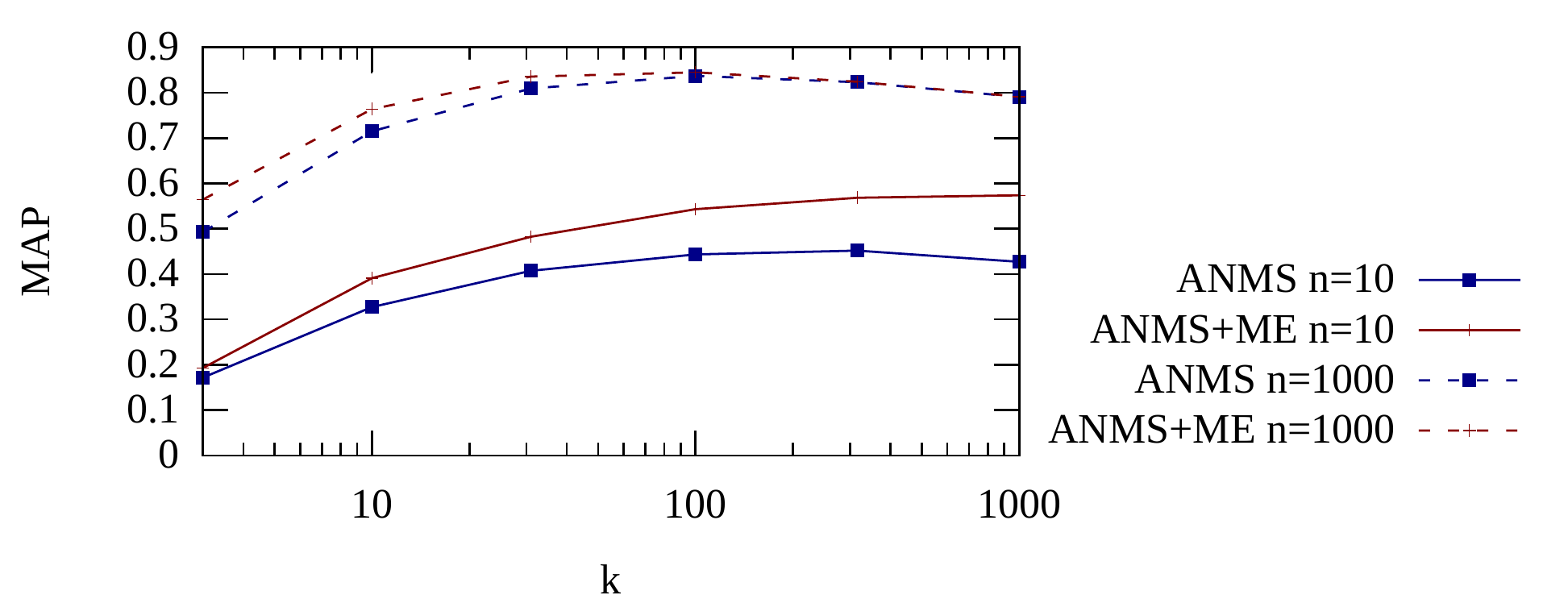}
  \vspace{-2em} 
  \caption{Performance for varying $k$ (Hessian-affine SIFT). Straight lines show the performance for 10 keypoints, dashed lines for 1000 keypoints.
  Equivalent approaches have equivalent colors.}
  \label{fig:hesaffSiftK}
  \vspace{-1em}
\end{figure}

\textbf{Value of $k$.} Not only the number of queried keypoints can be used to increase the number of seed hypotheses and therefore the matching
quality, but also $k$. While increasing $k$ comes at a lower query cost, it also produces hypotheses of lower quality. However, as it is well known
for example in the context of $k$NN classification, reasonable values for $k$ can increase the query performance. In this experiment we evaluate the effect of $k$ if the
number of features to be queried is fixed to a given number. For a large number of keypoints, a very high value of $k$ adds a lot of false positives
such that the MAP decreases \cite{JegDouSch11b}. On the other hand, if $k$ is too small, only a small amount of correct hypotheses is found
\cite{JegDouSch11b}. We reproduced this result with the ANMS ranker at 1000
keypoints (see Figure~\ref{fig:hesaffSiftK}), as here the MAP at $k=100$ is
highest compared to $k=10$ or $k=1000$.

What happens when we decrease the number of keypoints? As shown in
Figure~\ref{fig:hesaffSiftK}, if a large number of keypoints is queried
($n=1000$), then for all of the evaluated approaches a value of $k=100$
performed better than $k=1000$. So match expansion does not greatly affect the
optimal value of $k$ in this case. However, if only very few keypoints are used
for query processing (e.g. $n=10$), a large $k$ performed better with match
expansion. Without this additional step, performance decreased for large $k$
(however at a larger $k$ than at a higher number of keypoints queried), most
likely because the additional noise introduced could not be out-weighted by
the higher number of correct matches. This leads us to the following results:
The best way to increase query performance, which is well known, is to increase
the number of keypoints queried. In order to increase query performance however,
it is possible to decrease the number of keypoints queried. In this case, some
of the performance loss resulting from a lower number of keypoints can be
compensated by a large $k$ in combination with match expansion (and, at a lower
degree, even without expanding matches).

\textbf{Runtime.} The cost of the evaluated \textit{keypoint ranking} approaches is negligible for the random and response based ones, as these just
have to sort the query features, and about 7ms (SIFT) and 5ms (BinBoost) for the
ANMS ranker. For Hessian-affine SIFT on Oxford5k, scoring times (including ME)
were about 6ms for processing all $k$ results of a single $k$NN query ($k=100$, $n=100$), and therefore slightly lower than the runtimes of running a single $k$NN query which took about 7ms, at the possible gain of adding additional matches and a rough geometric check. The feature quantization needed for
match expansion took about 45ms for all features in a query image. For BinBoost features (including ME), the match expansion and scoring took less
than 4ms for processing a single $k$NN result. Runtimes increase with $k$, as more correspondences have to be expanded. The overall runtime for Hessian-Affine SIFT at
100 keypoints (ANMS+ME) was about 1.34s ($k=100$, $n=100$), while for binary
features it was higher (15s), as for this we used an exact,
though state-of-the-art, indexing technique.

Setting runtimes in relation to MAP, it is possible to beat an RND ranker
considering 100 keypoints with ANMS+ME considering 50 keypoints at a slightly
lower runtime 0.69s vs 0.75s and a higher MAP (see
Table~\ref{tab:k100SIFT_HESAFF_OXFORD5K_SC6_LOPQ_RSNL1_SQ8}, 0.741
vs. 0.698). For the holidays dataset runtimes of the random approach (RND) were
about 0.9s ($k=10$, $n=100$) and for ANMS+ME it was only approximately 0.6s
($k=10$, $n=50$) at a higher MAP.
The most time-consuming operation during match expansion is the search of
spatially close features. Therefore we think that the runtimes of match
expansion can be reduced significantly by optimizing this matching step, e.g. by
ordering features in a $k$d-tree which can be realized without additional space
overhead. This would also help on the Paris6k dataset where runtimes of the
random approach (RND) were about 0.8s ($k=100$, $n=100$) and for ANMS+ME
approximately 0.7s ($k=100$, $n=50$) at similar MAP.
Runtimes have been measured using only a single core. As each keypoint is
queried separately and match expansion is also achieved on a per-keypoint basis,
query processing can be easily extended to a multi-core setting.

For Oxford105k the runtime for match expansion was similar to Oxford5k: A single $k$NN query took slightly less than 8ms and expansion took about 7ms.

\textbf{Comparison to the State of the Art.} While the primary goal of this
research is not to increase the effectiveness of object recognition but rather
to reduce the number of features queried, let us still compare the results from
this paper to the state of the art in order to get insights into its
performance. We will compare to \cite{QinWenGoo13}, as the authors were using
the same Hessian-Affine SIFT features as we do and a similar recognition
pipeline involving Product Quantization. On the Oxford5k dataset the authors of
\cite{QinWenGoo13} achieved 0.78 points in MAP using 8 subquantizers for
indexing features using Product Quantization, i.e. a setting close to our
scenario. This corresponds to the performance we could achieve when querying 100
keypoints. However, using the pipeline from this chapter requires a larger
memory footprint; if we consider techniques with a memory footprint closer to ours,
\cite{QinWenGoo13} was able to achieve 0.83 points in MAP by approximating
features more accurately using 32 bytes per feature. On Paris6k, using 8
subquantizers, \cite{QinWenGoo13} achieved a MAP of 0.74, which is slightly
better than the performance of ANMS+ME at 100 keypoints (at a higher memory
footprint, performance of \cite{QinWenGoo13} was 0.76).
Concerning Oxford105k, a larger number of features (about 300) is needed to
achieve performance comparable to the state of the art (0.728
\cite{QinWenGoo13}). Finally note that the performance of our baseline is lower
for Holidays than the state of the art performance of $0.80$ ($0.84$
respectively) from \cite{QinWenGoo13}; this might be due to a different
quantization training set or related to their similarity measure, which is
however complementary to our approach and can be easily integrated into our
pipeline.

\section{Conclusion}
\label{sec:conslusion}
In this paper we evaluated an alternative pipeline for decreasing the runtimes of object recognition when $k$NN queries are used for the generation of
tentative correspondences instead of Bags of Visual Words. While the reduction of query features can have negative effects on query performance,
especially if the unmodified standard recognition pipeline is used, some simple modifications in the pipeline aiming at feature ranking and match
expansion can already produce good results at only a fraction of $k$NN queries. Some challenges however, remain. First, more techniques for
feature ranking will have to be investigated that provide good results for any type of keypoint descriptor and extractor. 
Additionally, improvements in the match expansion stage should aim at increasing efficiency and effectiveness. Due to the simple structure of the
pipeline used in this paper, these improvements can be easily integrated.

\bibliographystyle{abbrv}
\bibliography{abbrev,egbib}

\end{document}